\documentclass[runningheads]{llncs}
\usepackage{graphicx}

\usepackage{tikz}
\usepackage{comment}
\usepackage{amsmath,amssymb} 
\usepackage{color}

\usepackage[pagebackref=true,breaklinks=true,colorlinks,bookmarks=false]{hyperref}

\usepackage{xcolor}

\begin{document}
\pagestyle{headings}
\mainmatter
\def\ECCVSubNumber{1143}  

\title{People as Scene Probes} 

\titlerunning{People as Scene Probes}
%
\author{Yifan Wang\and
Brian L. Curless\and
Steven M. Seitz}
\authorrunning{Y. Wang et al.}
%
\institute{University of Washington, Seattle\\
\email{\{yifan1,curless,seitz\}@cs.washington.edu}}

\maketitle

\begin{abstract}

By analyzing the motion of people and other objects in a scene, we demonstrate how to infer depth, occlusion, lighting, and shadow information from video taken from a single camera viewpoint.  This information is then used to composite new objects into the same scene with a high degree of automation and realism.  In particular, when a user places a new object (2D cut-out) in the image, it is automatically rescaled, relit, occluded properly, and casts realistic shadows in the correct direction relative to the sun, and which conform properly to scene geometry.  We demonstrate results (best viewed in supplementary video) on a range of scenes and compare to alternative methods for depth estimation and shadow compositing.

\end{abstract}

\section{Introduction}
\label{sec:intro}

The presence of people in an image reveals much about scene structure.  
Each pedestrian effectively acts as a {\em depth probe}, whose image height is inversely proportional to distance.  Similarly, people act as {\em occlusion probes}, revealing which parts of the scene are in front of others, as they pass in front of or behind signs, trees, cars, fences, and other structures.  They also act as {\em light probes}, revealing both how the scene casts light on them (shade vs. sun), as well as how they cast shadows on the scene.  Taken together (and over many observations), these cues capture a great deal of information about the scene.

This paper presents techniques for inferring depth, occlusion, and lighting/shadow information from image sequences of a scene, through analysis of people (and other objects such as cars). A key property of our approach is that it is completely {\em passive} -- unlike prior use of light probes \cite{debevec2008rendering}, depth probes \cite{brostow1999motion}, or shadow probes \cite{chuang2003shadow} which require actively placing and/or moving special objects in the scene; we recover all of this information purely from the images themselves.

As an application, we focus on geometry- and lighting-aware image compositing, i.e., pasting new people or other objects into an image in a way that automatically accounts for depth, occlusions, lighting, and shadows.  For example, when you drag a segmented image of a person onto the image, it automatically resizes her to be larger in the foreground and smaller in the background.  Placing her on the stairs will correctly occlude the car behind, and the system will add a realistic shadow that bends over the stairs and onto the pavement below.  Dragging her behind the tree close to a building automatically adds partial occlusion from the branches in front, and the part of her you see is darker due to the shadow cast by the building.  All of these effects occur in real-time, as you move her to different locations in the image.

Our technical contributions include (1)~an automatic method for estimating occlusion maps from objects (e.g., people and cars) moving through a scene, (2)~a network that learns from a video of a scene how to cast shadows of newly inserted objects into that scene, including warping and occlusion of shadows due to scene geometry, and (3)~a method for implicitly estimating a ground plane based on heights of people observed throughout the scene without camera calibration or explicit depth estimation, enabling depth-dependent scaling of newly inserted objects based on their locations.  Combined with a technique for estimating approximate illumination -- modeled by just scaling whole-object brightness depending on placement in the image -- we demonstrate a novel system for compositing objects into the scene with plausible occlusions, object scaling, lighting, and cast shadows.  We show that components of the system outperform alternatives such as single-view depth estimation for occlusions and standard GANs for shadow insertion.  

We note that the method does have limitations: it works for a single scene at a time for which stationary camera video is available, assumes a single ground plane, and does not handle complex re-shading of inserted objects or their reflections off of specular surfaces in the scene, and the shadow generation works when inserting in-class objects that are observed in typical places in the scene (people on sidewalks and cars on streets, but not cars on sidewalks or, say, sharks placed anywhere). Despite these restrictions, we show that much can be learned about the geometry and lighting of a scene simply by observing the effects of everyday people and cars passing through, thus enabling new image compositing capabilities.

\section{Related Work}

\noindent {\bf Conditional Image Synthesis} \quad
Deep generative models can learn to synthesize images, including generative adversarial networks (GANs) \cite{Goodfellow2014} and variational autoencoders (VAE) \cite{Kingma2014a}. 
Conditional GANs \cite{brock2018large,mescheder2018training,mirza2014conditional,miyato2018cgans,odena2017conditional} are used to synthesize images given category labels. \cite{park2019semantic,wang2018high,isola2017image} focus on converting segmentation masks to photo-realistic images. They offer users an interactive GUI to draw their own segmentation masks and output a realistic image based on the given segmentation masks. However, these GANs do not leverage scene-specific geometry and lighting information, derived from many images. Our work embeds the scene's geometry and lighting into the GAN, to generate more realistic compositions. 

\noindent {\bf Image Composition} \quad
Lalonde \textit{et al.} \cite{lalonde2007photo} proposed a system for inserts new objects into existing photographs by querying a vast image-based object library.  Several authors have explored use of GANs to transform a foreground object to better match a background.  ST-GAN \cite{lin2018st} learns a homography of a foreground object conditioned on the background image. Compositional GAN \cite{azadi2018compositional} additionally learns the correct occlusion for the foreground object. SF-GAN \cite{zhan2019spatial} warps and adjusts the color, brightness, and styles of the foreground objects and embeds them into background images harmoniously. 
However, a realistic composition should also consider the foreground object's effect on the background (including shadows).

Some approaches aim to compose an object by rendering its appearance. \cite{hong2018learning} inserts an object into a scene based on a  specified location and bounding box. 
\cite{lee2018context} learns the joint distribution of the location and shape of an object conditioned on the semantic label map.  PS-GAN \cite{ouyang2018pedestrian} replaces a pedestrian's bounding box  by random noise and infills with a new pedestrian based on the surrounding context. \cite{lee2019inserting} blends the object with the background image in the bounding box, and learns a mapping to synthesize realistic images using both real and fake pairs. These works all train on images without hard shadows, and focus on person rather than shadow synthesis.  For example, they only synthesize an area around the person's bounding box (not including long shadows), and don't take into account shadow casting information from other images of the same scene.

\noindent {\bf Shadow Matting} \quad
Matting \cite{porter1984compositing} is an effective tool to handle shadows. 
\cite{chuang2003shadow} enables synthesizing correct cast shadows, by estimating a shadow displacement map obtained by {\em manually} waving a shadow-casting stick over every part of the scene.  Given an object to be composited, they can then synthesize correct shadows based on the object shape and shadow displacement map.
The related problem of shadow {\em removal} has also been explored by a number of authors, e.g., 
\cite{guo2012paired,zhang2015shadow,le2019shadow}.
We present the first shadow matting (synthesis) method that is completely {\em passive}, i.e., does not require manually waving a stick, but instead learns from the movement of objects (people and cars) in the scene itself.

\noindent {\bf Image Layering} \quad
Our work was inspired in part by \cite{brostow1999motion}, who first proposed using the motion of people (and other objects) to infer scene occlusions relationships.  As the technology in the 1990's was more limited, their approach required manual intervention and made a number of simplifying assumptions.
Less related to our work, but also worth mentioning is the use of layered representation for view synthesis, e.g. 
\cite{Shade1998,dhamo2019peeking,tulsiani2018layer,zhou2018stereo,srinivasan2019pushing}.
Like \cite{brostow1999motion}, our approach infers occlusion order purely from the movement of objects in the scene, but is entirely automated and leverages modern techniques for object detection and tracking.


\section{Problem Definition}
We start with a video of a scene, taken by a stationary camera.  As objects -- people, cars, bicycles -- pass through the scene, they occlude and are occluded by scene elements, pass into and out of shadowed regions, cast shadows into the scene, and, due to perspective, appear larger or smaller in an image depending on their position in the scene.  From the video sequence, we seek to extract occlusion layering, shadowing, and position-dependent scale to enable realistically compositing new objects (of similar classes) into the scene.


We design a fully automatic pipeline to tackle this problem. Our key idea 
is that the occurrence and motion of existing objects (aka {\em scene probes}) through the video is the primary cue for inferring properties of the scene. These properties include depth, occlusion ordering, lighting, and shadows. Unlike some related methods, our pipeline does not require active scanning for shadow matting~\cite{chuang2003shadow}, or manual annotation for layering~\cite{brostow1999motion}. Furthermore, our pipeline does not require the camera to be calibrated.

\section{Technical Approach}

\subsection{People as Occlusion Probes}
\label{sec:geometry}
Occlusion is key for realistic image composition. An inserted object should be occluded by the foreground and occlude the background properly.  Cars driving on the street are occluded by the trees on the sidewalk nearer to the viewer, and road signs occlude people walking behind them. We propose a method to estimate the occlusion order by {\em analyzing the occlusion relationships between people, other moving objects in the video, and static scene structures and objects}. Our method records the occlusion relationship between object and the scene to yield an occlusion map
$\tilde{z}(x,y)$, similar to a depth map, for determining which pixels of an object occlude or are occluded by the scene, depending on the location of the object. 
To make the problem tractable, we approximate the scene as a single ground plane, with moving objects and occluders represented as planar sprites (on vertical planes parallel to the image plane) that are in contact with the ground plane.  Based on this simplification, we can assume a monotonic relationship between object location and occlusion order; the closer the object, the lower in the image its ground contact occurs.


\subsubsection{Algorithm}
We first calculate a median image within a local temporal window of one second to serve as a background plate; we have found the one second window to work well for scenes that are not densely crowded, and with objects (especially people) moving at a natural pace. For each frame in this temporal window, we apply Mask-RCNN~\cite{he2017mask} to estimate segmentation masks for people, cars, bikes, trucks, buses, and related categories.  For each individual object $O_i$, Mask-RCNN returns a binary mask $M_i$, and we record the lowest point $y_i$ of the mask.  We refine this mask to avoid accidental inclusion of background pixels: each pixel in $M_i$ whose color difference with the median image is greater than a threshold is assigned to refined mask $M'_i$.

Now we construct the occlusion map.  We set the image origin $(x,y) = (0,0)$ at the lower left corner of the image.  The key idea is that if an object $O_i$ with bottom pixel $y_i$ occludes a background pixel, then another object $O_j$ with $y_j < y_i$ is likely to be closer to the camera and would then also occlude this pixel. We initialize the occlusion map with $\tilde{z}(x,y) = -1$ at all pixels and then iteratively update the map for each object $O_i$:
\begin{equation}
    \tilde{z}(x,y) = 
      \begin{cases}
         y_i,& \text{if } (x,y) \in M'_i \text{ and } y_i > \tilde{z}(x,y)\\
         \tilde{z}(x,y), & \text{otherwise}.
      \end{cases}
\end{equation}
To create a new composite, we initialize image $I_{\rm comp}$ with one of the median images.  For a new object $O_j$ (e.g., cropped from another photo) with mask $M_j$ and bottom coordinate $y_j$, we update $I_{\rm comp}$:
\begin{equation}
    I_{\rm comp}(x,y) = 
      \begin{cases}
         O_j(x,y), & \text{if } (x,y) \in M_j \text{ and } y_j < \tilde{z}(x,y)\\
         I_{\rm comp}(x,y), & \text{otherwise}.
      \end{cases}
\end{equation}
where $O_j(x,y)$ is the color of the object at a given pixel $(x,y)$.  Note that this composite image lacks shadows cast by $O_j$.  Further, if $O_j$ is inserted into an area that is itself in shadow, then $O_j$ should be darkened before compositing.  We discuss these shadowing effects in the next section.




\subsection{People as Light Probes}
\label{sec:scene-shadow}
People appear brighter in direct sunlight and darker in shadow.  Hence, we can potentially use people to {\em probe} lighting variation in different parts of a scene.  Based on this cue, we compute a lighting map that enables automatically adjusting overall brightness of new objects as a function of position in the image.
We do not attempt to recover an environment map to relight objects, or to cast complex/partial shadows on objects (areas for future work).  Instead, we simply estimate a  darkening/lightening factor to apply to each object depending on its location in the scene, approximating the effect of the object being in shadow or in open illumination.  We call this factor, stored at each pixel, the lighting map $L(x,y)$.  This lighting map is a {\em spatially varying} illumination map across the image, whereas the prior work~\cite{georgoulis2017around,park2020seeing,hold2019deep,hold2017deep} generally solves for a single {\em directionally varying} illumination model for the entire scene.  From the input video, we observe that people walking in well-lit areas tend to have higher pixel intensity than people in shadowed areas.  We further assume there is no correlation between the color of people's clothing and where they appear in the image; e.g., people wearing red do not walk along different paths than those wearing blue. Given these conditions, we estimate the lighting map from statistics of overall changes in object colors as they move through the scene.  Note that this lighting map is a combination of average illumination and reflection from the surface; it does not give absolute brightness of illumination, but gives a measure of relative illumination for different parts of the scene.


\subsubsection{Algorithm}
Starting with the detected objects $\{O_i\}$ and associated masks $\{M'_i\}$ described in Section~\ref{sec:geometry}, we compute the mean color $C_i$ per object across all pixels in its mask.  The lighting map is then the average of the $C_i$ that cover a given pixel, i.e.:
\begin{equation}
    L(x,y) = \frac{1}{|\{i \mid (x,y) \in M'_i\}|}\sum_{i \mid (x,y) \in M'_i} C_i
\end{equation}
When compositing a new object, $O_j$ with mask $M_j$ into the background plate, we first compute the average lighting $L_j$ for the pixels covered by $M_j$:
\begin{equation}
    L_j = \frac{1}{|\{(x,y) \in M_j\}|}\sum_{(x,y) \in M_j} L(x,y)
\end{equation}
and apply this color factor component-wise to all of the colors in $O_j$. As noted above, this lighting factor makes the most sense as a relative measure.  Thus, when compositing a new object into the scene in our application scenario, the user would first set the brightness of the object at a given point in the scene (with the lighting multiplied in), and can then move the object to different parts of the scene with plausible changes to the brightness then occurring automatically.



\begin{figure}[t]
    \centering
    \includegraphics[width=\linewidth]{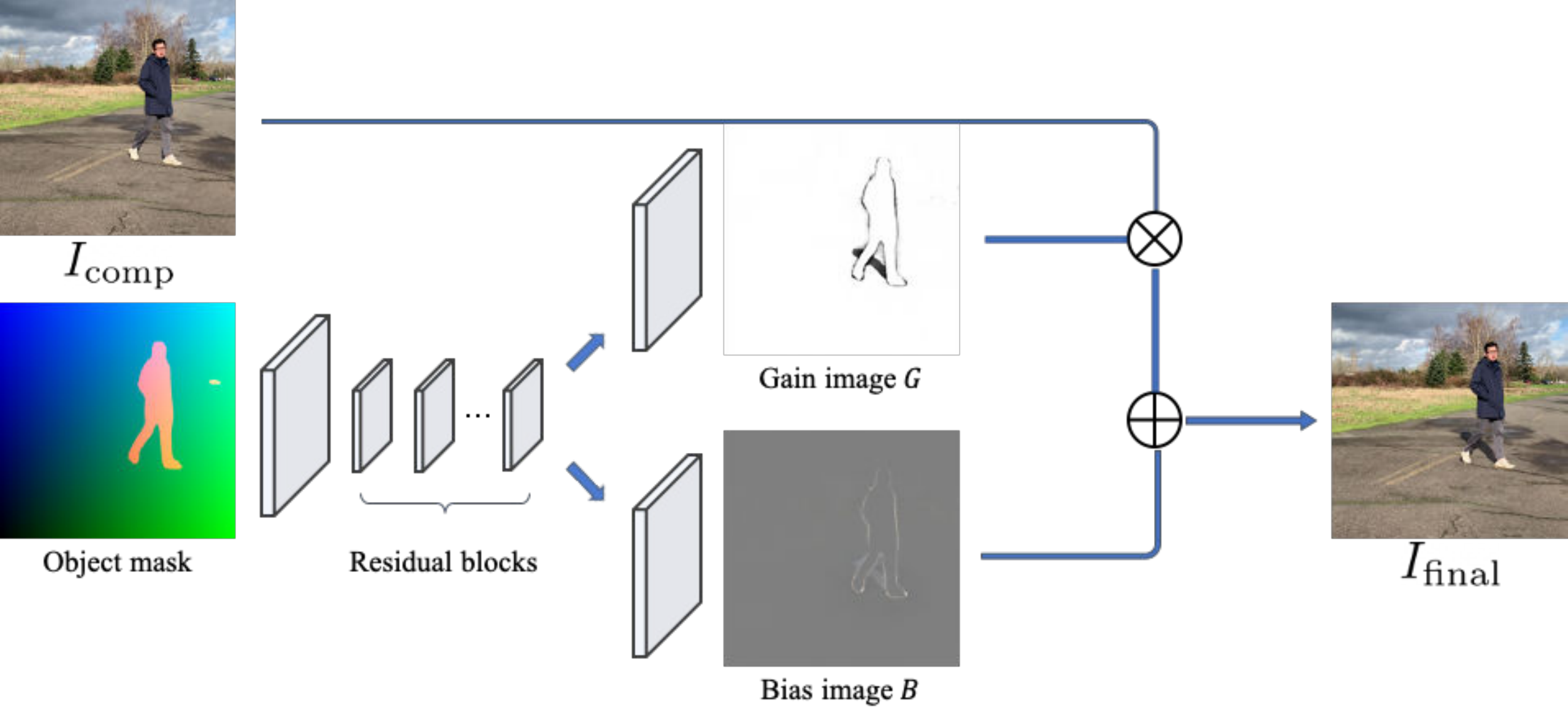}
    \caption{The generator network takes the object mask, $x$, and $y$ coordinates as input (visualized here in red, green, and blue channels) and outputs a scalar gain image $G$ and color bias image $B$ (mid-gray corresponds to zero bias as shown).  Given the shadow-free, composite image $I_{\rm comp}$, we synthesize the final image $I_{\rm final} = G \cdot I_{\rm comp} + B$.}
    \label{fig:network}
\end{figure}

\subsection{People as Shadow Probes}
\label{sec:cast-shadow}
Shadows are one of the most interesting and complex ways that moving objects interact with a scene.  Predicting shadows is challenging, as their shapes and locations depend on the position of the sun in the sky, the weather, and the geometry of both the object casting the shadow and the scene receiving it.  Furthermore, unlike other lighting effects, shadows are not additive, as a surface already in shadow does not darken further when a second shadow is cast on it from the same light source.
We propose using observations of objects passing through the scene to recover these shadowing effects, using a deep network -- a pix2pix~\cite{isola2017image} GAN with improved losses~\cite{wang2018high} -- trained on the given scene to learn how objects cast shadows depending on their shapes and locations in scene.  Further, since the discriminator encourages generation of realistic images, the network also tends to improve jagged silhouettes.


\subsubsection{Algorithm}

A natural choice of generator would take as input a shadow-free, composite image $I_{\rm comp}$ and directly output an image with shadows.  In our experience, such a network does not produce high quality shadows, typically blurring them out and sometimes adding unwanted color patterns.  Instead, we use the object masks of inserted objects as input, which are stronger indicators of cast shadow shapes.  Inspired by \cite{liu2018intriguing}, we concatenate an image channel comprised of just the per-pixel $x$-coordinate, and another channel with just the per-pixel $y$-coordinate; we found that adding these channels was key to learning shadows that varied depending on the placement of the object, e.g., to ensure the shadow warped across surfaces or was correctly occluded when moving the object around.  As in Figure~\ref{fig:network}, we feed this $x$-$y$ augmented object mask through a deep residual convolutional neural network to generate a scalar gain image $G$ and color bias image $B$, similar to the formulation in \cite{le2019shadow,le2018a+}.  The final image is then $I_{\rm final} = G \cdot I_{\rm comp} + B$.  We found that having the generator produce $I_{\rm final}$ directly resulted in repetitive pattern artifacts that were alleviated by indirectly generating the result through bias and gain images.

For training, we take each input image $I$ and follow the procedure in Section~\ref{sec:geometry} to extract objects $\{O_i\}$ and masks $\{M'_i\}$ from an image and then composite the objects directly back onto the local median image to create the shadow-free image $I_{\rm comp}$.  The resulting $I_{\rm final}$, paired with the ground truth $I$, can then be used to supervise training of the generator and discriminator, following the method described in~\cite{wang2018high}.

\subsection{People as Depth Probes}
\label{sec:plane}
The size of a person (or other object) in an image is inversely proportional to depth.  Hence, the presence of people and their motion through a scene provides a strong depth cue.  Using this cue, we can infer how composited people should be resized as a function of placement in the scene.
We propose a method to estimate how the scale of an object should vary across an image without directly estimating scene depth or camera focal length, but based instead on the sizes of people at different positions in the scene.  Our
problem is related to \cite{bose2003ground} who rectify a planar image by tracking moving objects, although they require constant velocity assumptions, which we avoid.  \cite{criminisi2000single} determines the height of a person using a set of parallel planes and a reference direction, which we do not require.  We make two assumptions: (1) the ground (on which people walk) can be approximated by a single plane, and (2) all the people in the video are roughly the same height. While the second assumption is not strictly true, it facilitates scale estimation, essentially treating individual height differences among people as Gaussian noise, as in \cite{Hoiem2008}, and solving via least squares.


\subsubsection{Algorithm}
According to our first assumption, all ground plane points $(X,Y,Z)$ in the world coordinate should fit a plane equation: 
\begin{equation}
    a X + b Y + c Z = 1
    \label{eqn:world_plane}
\end{equation}

Under the second assumption, all people are roughly the same height $H$ in world coordinates. Under perspective projection, we have:
\begin{equation}
    x = X \cdot \frac{f}{Z}, y = Y \cdot \frac{f}{Z}, h = H \cdot \frac{f}{Z}
\end{equation}
where $f$ is the focal length of the camera. Multiplying both sides of Equation~\ref{eqn:world_plane} by $H \cdot \frac{f}{Z}$, we arrive at a linear relation between pixel coordinates and height:
\begin{equation}
    a' x + b' y + c' = h
    \label{eqn:camera_plane}
\end{equation}
where $a', b', c'$ are constants. Because people in the scene are grounded, Equation~\ref{eqn:camera_plane} suggests that any person's bottom middle point $(x_i, y_i)$ and her height $h_i$ follow this linear relationship.

Given the input video, we use the same segmentation network as in Section~\ref{sec:geometry} to segment out all the people in the video. For each person in the video, we record her height $h_i$ and bottom middle point $(x_i, y_i)$ in camera coordinates. After collecting all the $(x_i, y_i)$ and $h_i$ from the image segmentation network, we use the least squares method to solve for the $(a', b', c')$ in Equation~\ref{eqn:camera_plane}.

When inserting a new object into the scene at $(x_j, y_j)$, we apply Equation~\ref{eqn:camera_plane} to estimate height $h_j$. The inserted object will then be resized accordingly and translated to $(x_j, y_j)$. In our application, if the user requires a different height for an inserted object, then she can simply place the object and rescale as desired, and the system will then apply this rescaling factor on top of the height factor from Equation~\ref{eqn:camera_plane} when moving the object around the scene.


\subsection{Implementation Details}
We use Mask-RCNN \cite{he2017mask} as the instance segmentation network.  Inspired by \cite{johnson2016perceptual}, our shadow network uses a deep residual generator.  The generator has 5 residual blocks, followed by two different transposed convolution layers to output the bias and gain maps.  The loss function is the same as in \cite{wang2018high}.  We use ADAM with an initial learning rate of 1e-4, and decays linearly after 25 epochs to optimize the objectives.  More details can be found in supplementary.

\section{Results and Evaluation}
In this section, we first introduce our collected datasets, and then evaluate our entire pipeline, including individual components.

\begin{figure}[t!]
    \centering
    \includegraphics[width=.9\linewidth]{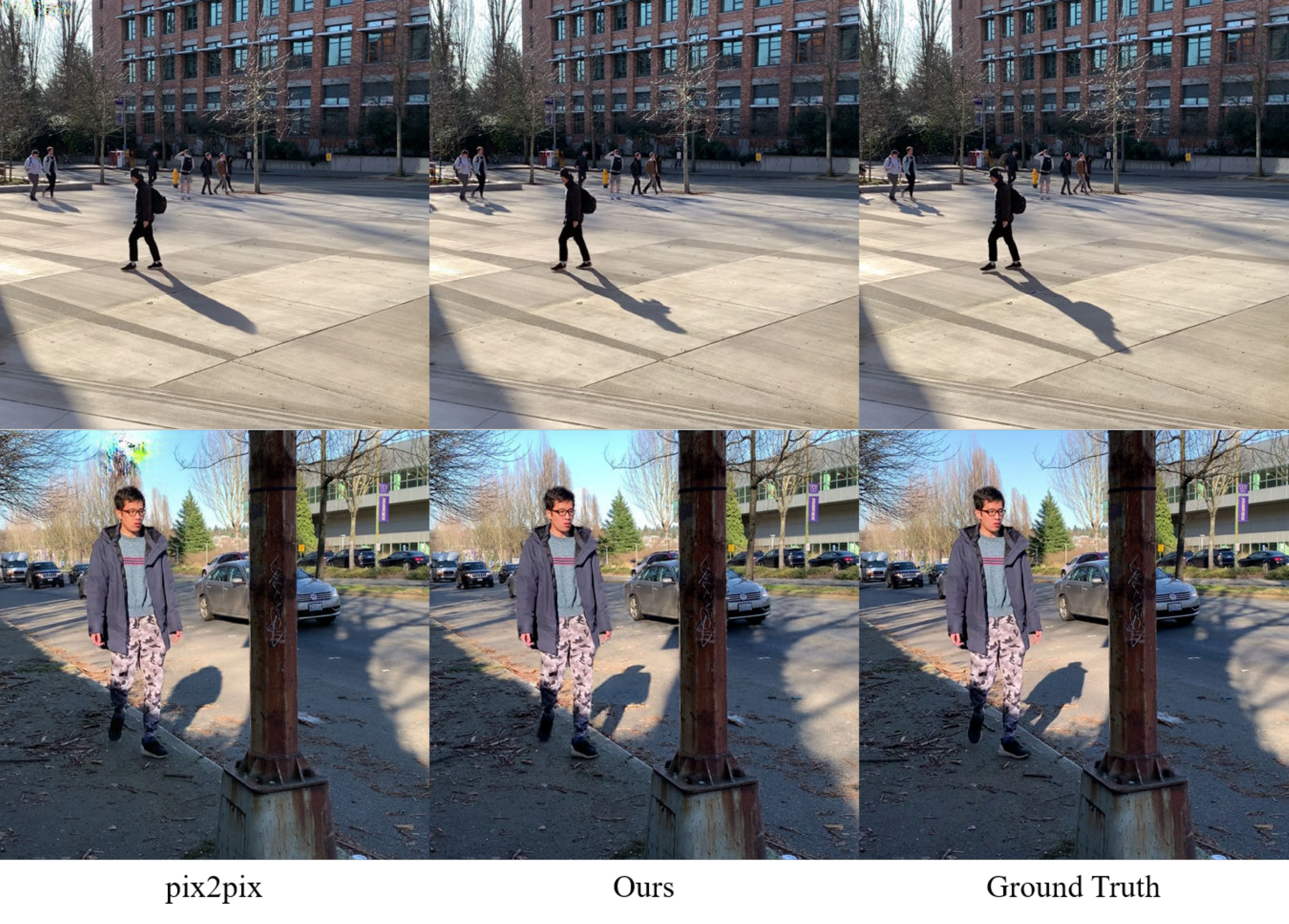}
    \caption{Shadow synthesis results on the test set. The images synthesized by a conventional pix2pix approach~\cite{wang2018high}~(left) lack details inferred by our network~(middle) which more closely resembles ground truth~(right).  In addition, the pix2pix method injects color patterns above the inserted person in the bottom row.  Note that both networks learn not to further darken existing shadows (bottom row, sidewalk near the feet).}
    \label{fig:shadow_results}
\end{figure}

\begin{figure}[ht!]
    \centering
    \includegraphics[width=.9\linewidth]{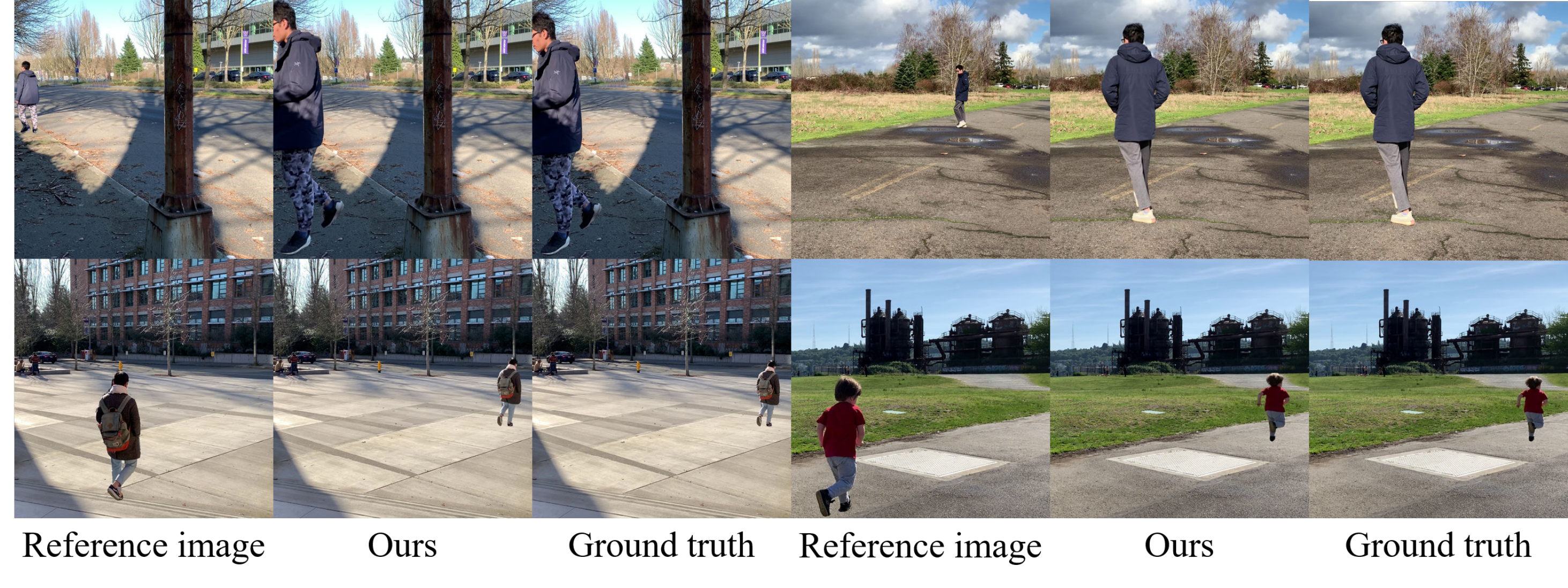}
    \caption{Ground plane estimation for height scaling. Given the person's reference height (taken from person in reference image), our algorithm accurately estimates the height in a different location (middle composite in each set). The difference between our estimate and the ground truth is small.  (Each of these images is a composite before adding shadows.)}
    \label{fig:height_results}
\end{figure}

\subsection{Data}
\label{sec:data}
We collected 11 videos with an iPhone camera on a monopod. These videos cover a range of scenes including city streets, parks, plazas, beaches, etc, under lighting conditions from clear sky to cloudy day. The videos are $25$ minutes long on average, during which the ambient lighting changes little. We center-crop to $800 \times 800$ at $15$ fps in training. We use the first $95\%$ of the video for training and the last $5\%$ for test.

    

\begin{figure}[p!]
    \centering
    \includegraphics[width=.92\linewidth]{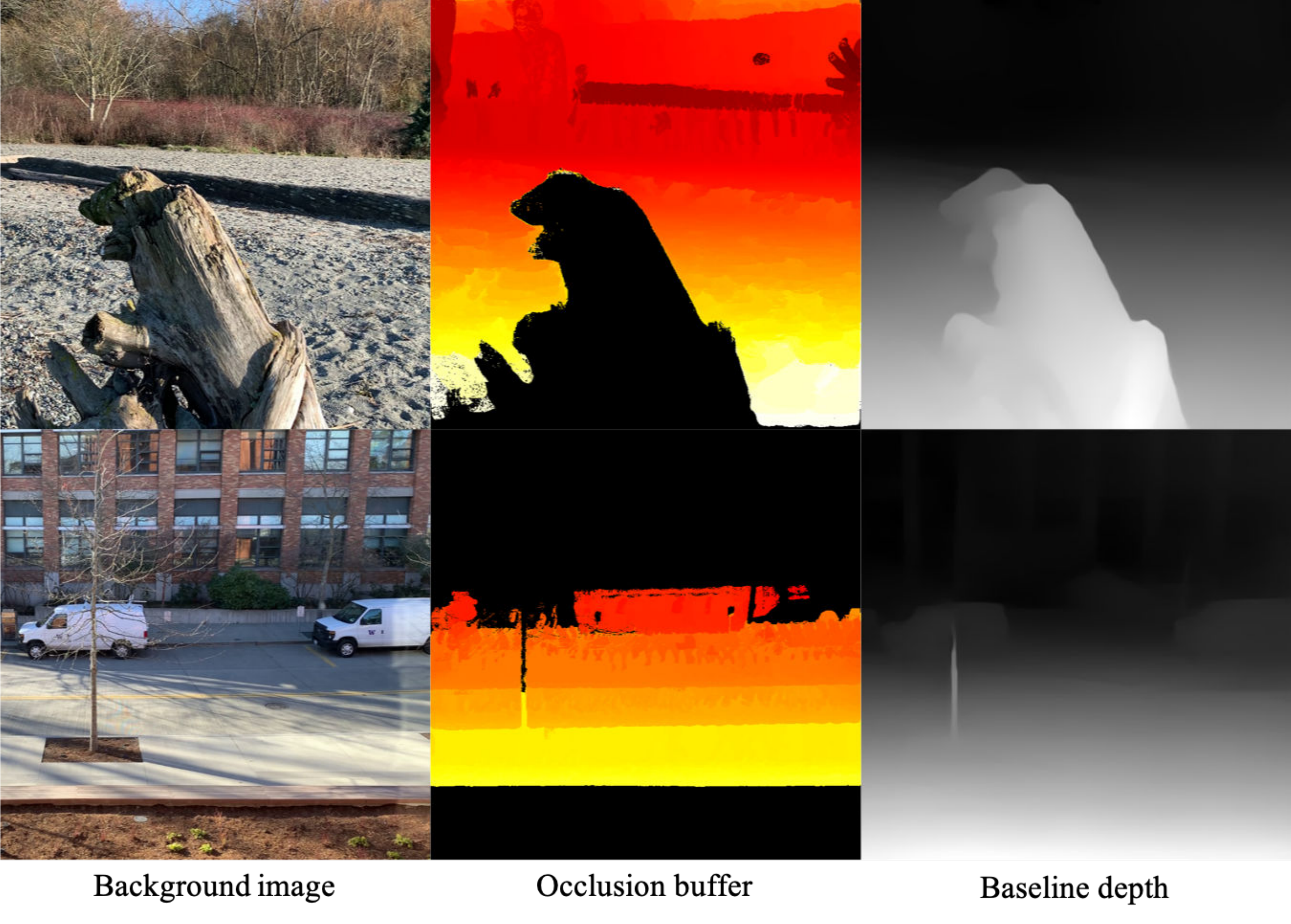}
    \caption{From left to right: the background image, our estimated occlusion buffer, and the depth predicted by~\cite{lasinger2019towards}. In our occlusion buffer, black pixels are never occluded. Pixels toward yellow were occluded only by smaller $y$-value objects, and pixels toward red were occluded by larger $y$-value objects.  Some quantization and noise arises in our occlusion map due, respectively, to common object placements (cars in the road) and object mask errors (arising from object/background color similarity at a given pixel).}
    \label{fig:occlusion_depth}
\end{figure}

\begin{figure}[htbp!]
    \centering
    \includegraphics[width=.92\linewidth]{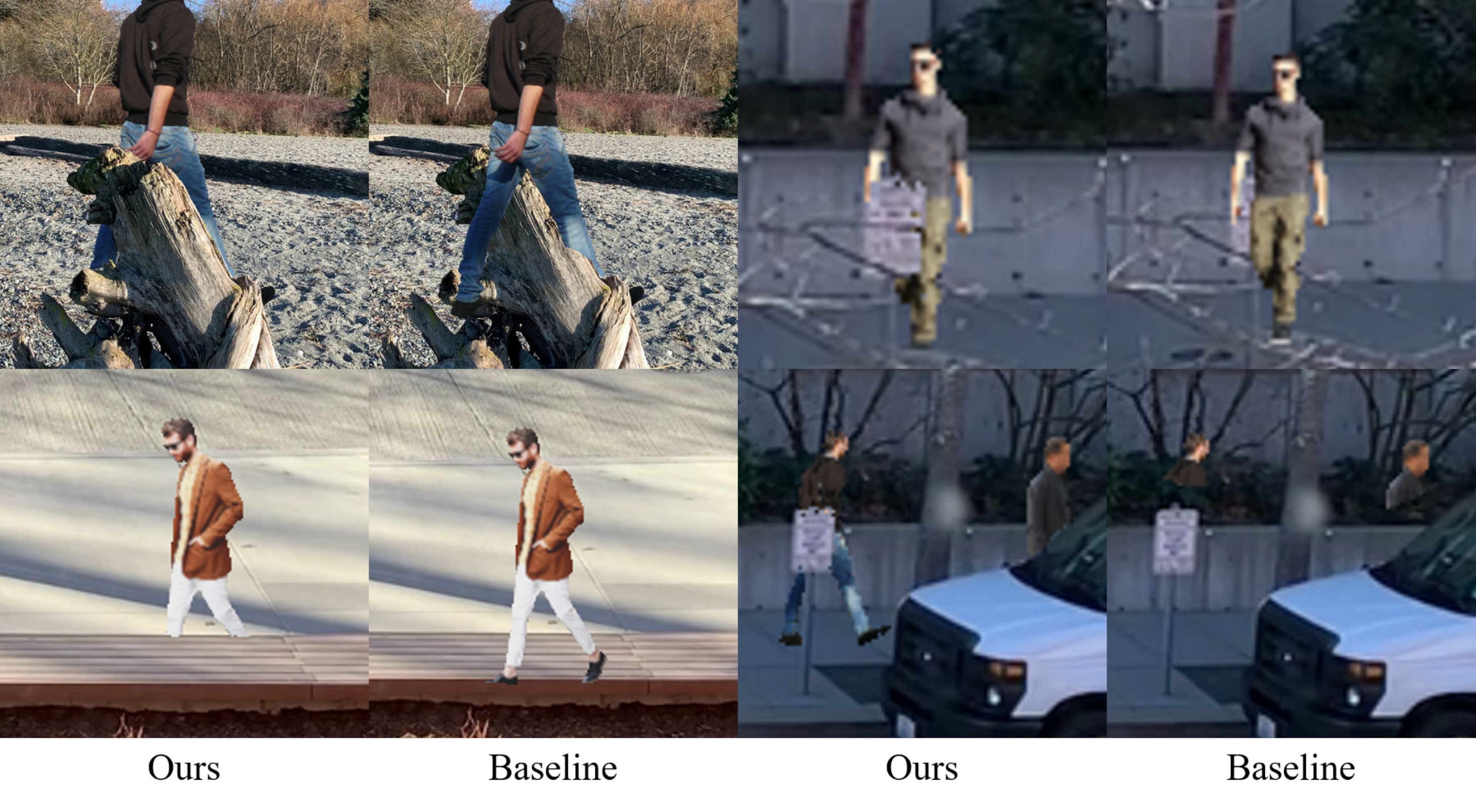}
    \caption{Qualitative results on occlusion order estimation. People are reasonably well-occluded by foreground objects using our occlusion buffer, while errors in the depth map approach~\cite{lasinger2019towards} (e.g., sign not reconstructed, bench reconstructed at ground level) result in incorrect occlusions.  (Each of these images is a composite before adding shadows.)}
    \label{fig:geometry_results}
\end{figure}

\subsection{Occlusion Probing}
\label{sec:eval_geometry}
Following Section~\ref{sec:geometry}, we generated occlusion maps for each scene; two of them are illustrated in Figure~\ref{fig:occlusion_depth} in yellow to red pseudocolor.  The quantization in colors corresponds to how objects moved through the scene; e.g., the two tones in the street correspond to the restricted placement of cars, which are generally centered in one of two lanes.  The black regions correspond to pixels that were never observed to be covered by an object; these are treated as never-to-be-occluded during compositing.  As a baseline, we also constructed depth maps using the state-of-the-art, depth-from-single-image MiDaS network~\cite{lasinger2019towards}.  MiDaS produces visually pleasing depth maps, but misses details that are crucial for compositing, such as the street signs toward the back of the scene in the second row of Figure~\ref{fig:occlusion_depth}.  

Figure~\ref{fig:geometry_results} shows several (shadow-free) composites.  For our method, we simply place an object (such as a person, scaled by the method in Section~\ref{sec:plane}) into the scene, and it is correctly occluded by foreground elements such as trees, benches, and signs.  For the baseline method, the depth of the element must be determined somehow.  Analogous to our plane estimation method for height prediction, we fit a plane to the scene points at the bottoms of person detections and then placed new elements at the depths found at the bottom-middle pixel of each inserted element.  In a number of cases, elements inserted into the MiDaS depth map were not correctly occluded as shown in the figure, due to erroneous depth estimates and the difficulty of placing the element at a meaningful depth given the reconstruction.

\subsection{Shadow Probing}
\label{sec:eval_light}
We trained our shadow estimation network (Section~\ref{sec:cast-shadow}) on each scene separately, i.e, one network per scene.  On average, each scene had $17,000$ images for training with $900$ images held out for testing.  Figure~\ref{fig:shadow_results} shows example results for shadow estimation using (1)~a baseline pix2pix-style method~\cite{wang2018high} that takes a shadow-free image and directly generates a shadowed image and (2)~our method that takes an $x$-$y$-mask image and produces bias and gain maps which are applied to the shadow-free image.  Both networks had similar capacity (with $5$ residual blocks).  In this case, we also had ground truth, as we could segment out a person from one of the test images and copy them into the median background image for processing.  The conventional pix2pix network tends to produce ``blobbier'' shadows when compared to the more structured shadows produced by our method, which is generally more similar to ground truth.

\subsection{Depth (Ground Plane) Probing}
\label{sec:eval_plane}
For each input video, we predict the plane parameters from the training images using the method described in Section~\ref{sec:plane}. When inserting a new object into the scene, we apply Equation~\ref{eqn:camera_plane} with regressed plane parameters to get its estimated height. We then resize it based on the estimated height, and copy-paste it onto the background frame.

To numerically evaluate the accuracy of the plane estimation as height predictor, we use it to measure relative, rather than absolute, height variation across the image.  This measure factors out errors due to, e.g., children not being of average adult height as the absolute model would predict.  In particular, we take one image as reference and another as a test image with the same person at two different positions in the images. Suppose Equation~\ref{eqn:camera_plane} predicts height $h$ in the reference image, but the actual height of the object is observed to be $\hat{h}$.  The prediction ratio is then $r = \hat{h}/h$.  For the same person in the test image, we then predict the new height $h'$ again using Equation~\ref{eqn:camera_plane} and rescale the extracted person by $r \cdot h'$ before compositing again.  We compared this rescaled height to the actual height of the person in the test image and found that on a small set of selected reference/test image pairs, the estimates were within $3\%$ of ground truth.  Figure~\ref{fig:height_results} illustrates this accuracy qualitatively.

Note that without relative height prediction, i.e., instead using Equation~\ref{eqn:camera_plane} to predict absolute heights, the height prediction error was $13.28\%$, reasonable enough for inserting an adult of near-average height, though of course more objectionable when inserting, say, a young child.  In our demo application, we allow the user to change the initial height of the inserted element.

\begin{figure}[t!]
    \centering
    \includegraphics[width=\linewidth]{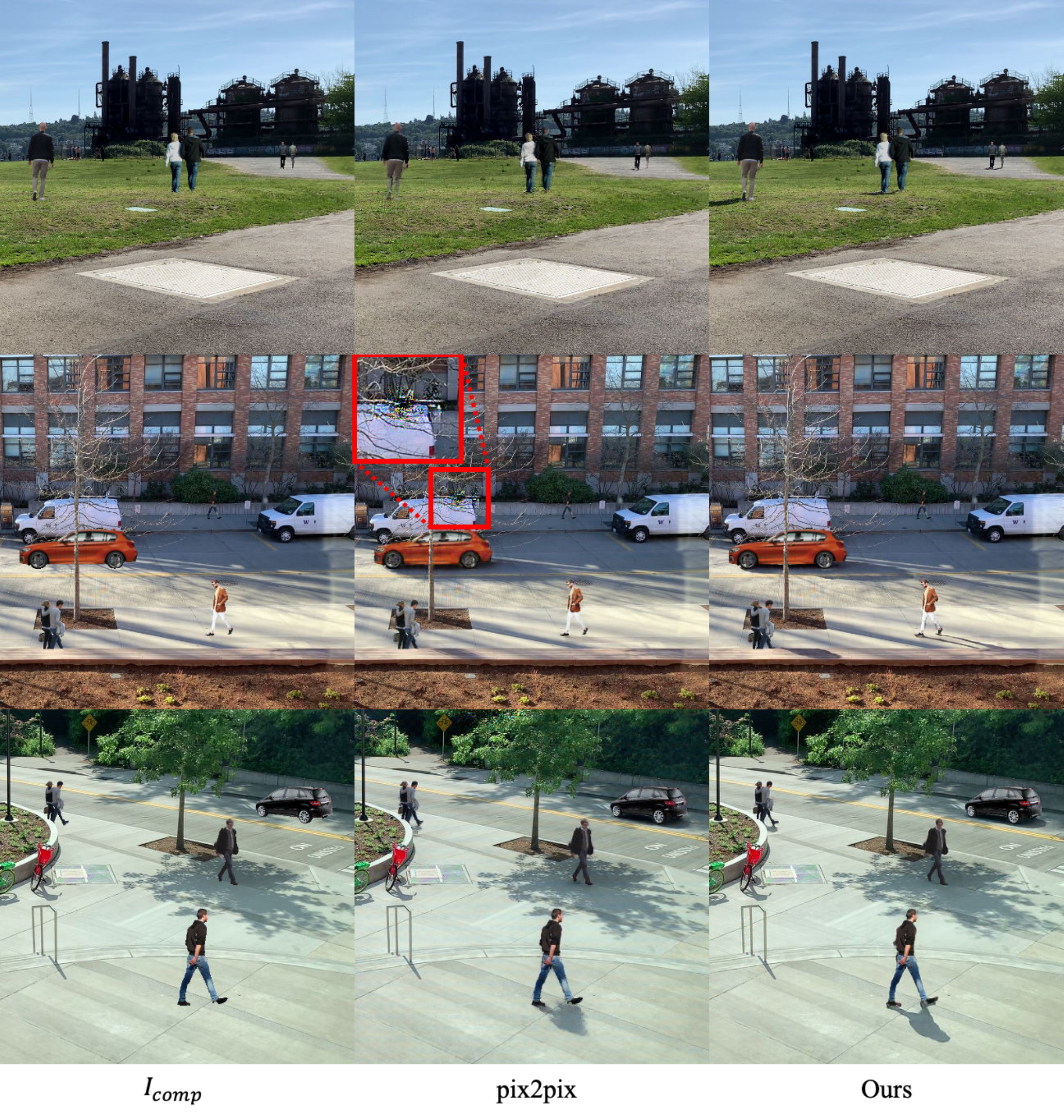}
    \caption{Inserting new objects. From left to right: the shadow-free composite image $I_{\rm comp}$, image synthesized by pix2pix~\cite{wang2018high}, and image synthesized by our shadow network. After applying our automatic scaling, lighting, and occlusion handling, we can see that our shadow network performs better than pix2pix on the final shadow compositing step; pix2pix generates less-detailed shadows and sometimes none at all and can introduce new artifacts (e.g., color patterning in the branches above the inserted car in the second row).  Our method warps detailed shadows over surfaces.  Both methods successfully avoid double-darkening of shadows (man in tree shadow in bottom row).}
    \label{fig:insert_results}
\end{figure}

\begin{figure}[t]
    \centering
    \includegraphics[width=\linewidth]{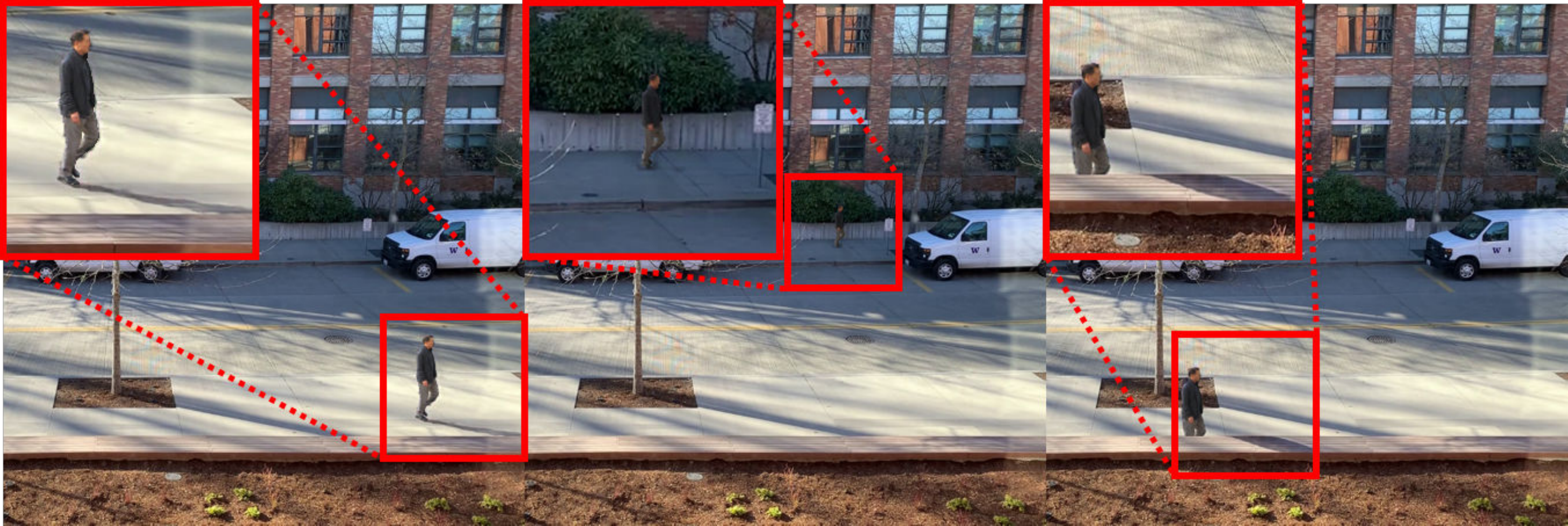}
    \caption{Our system adjusts the height, illumination, occlusion, and shadow for the same person based where he is placed in the image.}
    \label{fig:ui}
\end{figure}
\begin{figure}[t]
    \centering
    \begin{minipage}{.32\textwidth}
        \centering
        \includegraphics[width=\linewidth]{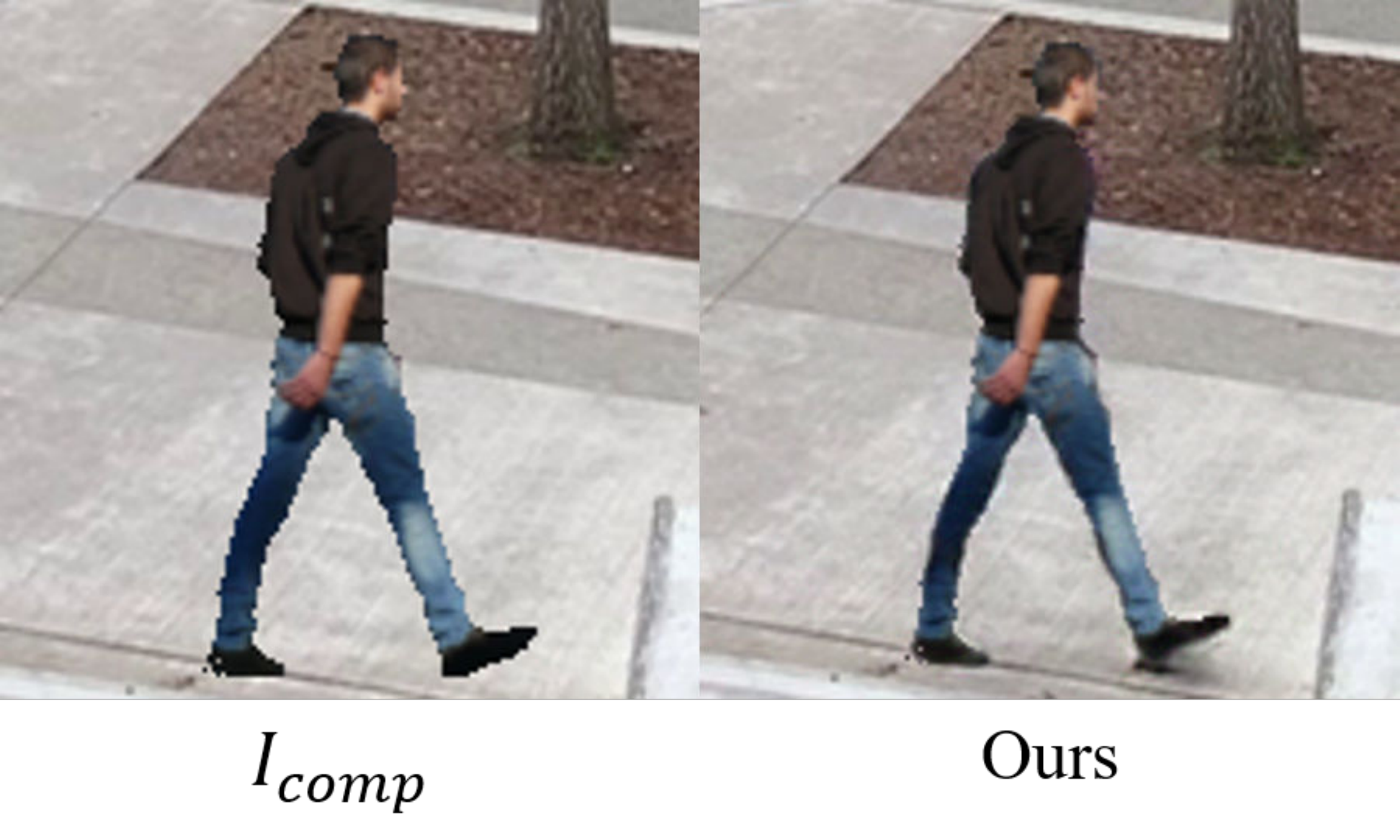}
        \caption{Our shadow network reduces aliasing artifacts in $I_{\rm comp}$, smoothing out the boundary.}
        \label{fig:alias}
    \end{minipage} \quad
    \begin{minipage}{.64\textwidth}
        \includegraphics[width=\linewidth]{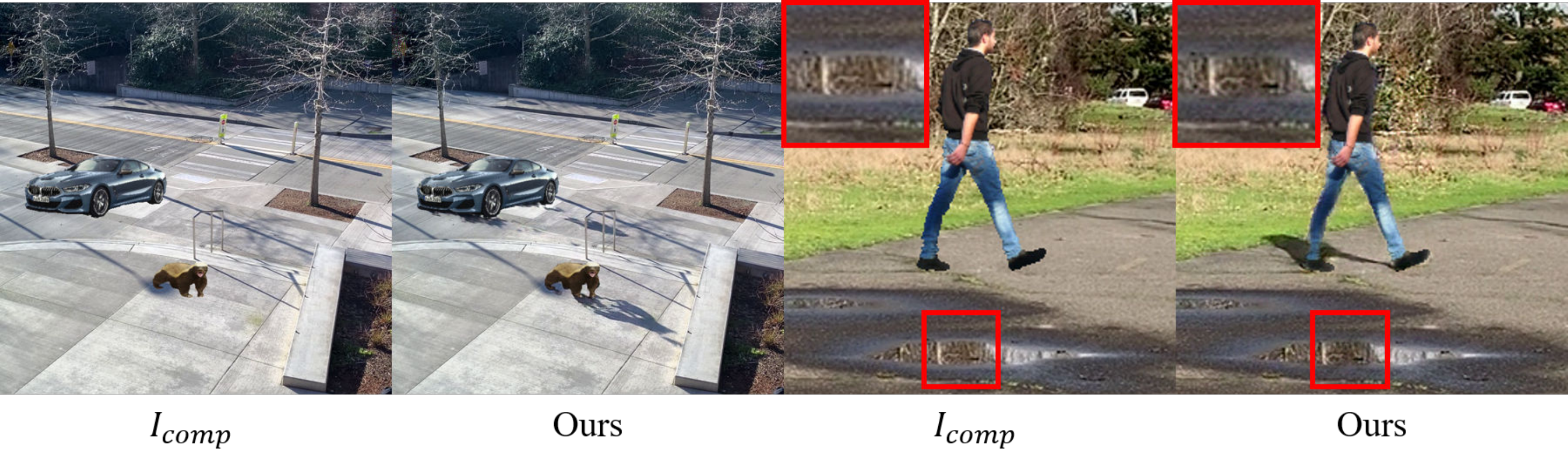}
        \caption{Our shadow network does not perform well on misplaced objects or objects not observed in the training set (left). On the right, the person should be reflected in the puddle, but is not.}
        \label{fig:failure}
    \end{minipage}
\end{figure}

\subsection{Inserting New Objects}
\label{sec:inserting}
We have developed an interactive image compositing application (see suppl. video) that leverages the elements of our system.  With this tool, a user can take an object (e.g., a person or car) downloaded from the internet or segmented from another image and drag it to a desired location in one of our captured background plates.  The system both resizes the object and adjusts its brightness depending on where it is placed, applies our occlusion map, and synthesizes any cast shadows.  We do not fully re-light or re-orient objects, so we rely on the user to select objects with generally compatible lighting and orientation.  We provide the user with the option of adjusting the height and overall brightness of the object after initially placing it, but afterward the height and brightness are updated automatically as the user moves the object around in the scene.

Figure~\ref{fig:insert_results} shows several results for object insertion using our tool.  Here we again compare to using the pix2pix variant just for the final shadow generation step.  Our method produces crisper shadows whereas the pix2pix method sometimes produces no shadow at all, generalizing poorly on some of the newly inserted objects and sometimes injecting undesirable color patterns.  We conducted human studies to quantify the realism of the synthesized shadows and found that our synthesized images were preferred by users $70.0\%$ of the time to baseline pix2pix, demonstrating a clear advantage for novel composites.  Details of the study appear in the supplementary material.

In Figure~\ref{fig:ui}, we demonstrate the effect of moving a person around in the scene.  Note how the brightness and height changes when moved from the lit front area to the back shadowed area, and how the person is occluded by the foreground bench with shadow wrapped over the bench when bringing the person closer to the camera.  We also show in Figure~\ref{fig:alias} how the shadow network reduces aliasing artifacts arising from the object's binary mask when initially inserted.

\section{Conclusion}
In this paper, we have introduced a fully automatic pipeline for inferring depth, occlusion, and lighting/shadow information from image sequences of a scene.  The central contributions of this work are recognizing that so much information can be extracted just by using people (and other objects such as cars) as scene probes to passively scan the scene. We show that the inferred depth, occlusion ordering, lighting, and shadows are plausible, with the occlusion layering and shadow casting methods outperforming single-image depth estimation and traditional pix2pix shadow synthesis baselines. Further, we show results using a tool for image compositing based on our synthesis pipeline.

As noted earlier, our method is not without limitations, requiring a single scene video as input, assumes a ground plane, does not model advanced shading effects, and cannot composite arbitrary objects at arbitrary locations.  Figure~\ref{fig:failure}, e.g., highlights two failure modes; objects either placed in unusual locations or objects in previously unseen categories do not result in plausible shadows, and reflections of objects off of reflective surfaces in the scene are not handled correctly.  These limitations point to a number of fruitful areas for future work.

\section*{Acknowledgement}
This work was supported by the UW Reality Lab, Facebook, Google, and Futurewei.

%
%
\bibliographystyle{splncs04}
\bibliography{shadow}

\clearpage
\section*{Supplementary}

\section*{A. Implementation Details}
We use Mask-RCNN \cite{he2017mask} with ResNet-152 backbone as the instance segmentation network.  For occlusion probes in Section~\ref{sec:geometry}, we use a local median window of one second to compute the local background frame.  The confidence threshold for Mask-RCNN is set to $0.75$ for both occlusion probes in Section~\ref{sec:geometry} and light probes Section~\ref{sec:scene-shadow}.  We segment out person, bicycle, car, motorcycle, bus, truck, backpack, umbrella, handbag, tie and suitcase for a complete set of moving objects.  For depth probes in Section~\ref{sec:plane}, we only segment out person and use a confidence threshold of $0.9$.  The high threshold usually gives a complete segmentation of a person's full body, which reduce the noise in estimating the ground plane parameters in Equation~\ref{eqn:camera_plane}.  To generate the data for shadow probes in Section~\ref{sec:cast-shadow}, person, bicycle, car, motorcycle, bus, truck, backpack, umbrella, handbag, tie and suitcase are segmented out with a confidence threshold of $0.8$.  The local median window is set to be $50$ frames for the shadow-free composite image $I_{\rm comp}$.

Inspired by \cite{johnson2016perceptual}, our shadow network uses a deep residual generator. It has 5 residual blocks instead of 9 in \cite{wang2018high}, because predicting the bias and gain map is an easier task than synthesizing the whole frame. Two different transposed convolution layers follow the decoder to output the bias and gain maps. The loss function is the same as in \cite{wang2018high}, i.e., two multi-scale discriminators with LSGAN loss, a feature matching loss, and a VGG perceptual loss. The initial learning rate is set to 1e-4, and decays linearly after 25 epochs. It decays to $0$ in another 25 epochs. The batch size is $4$. We train our network on four Nvidia RTX 2080 Ti GPUs, and each iteration takes about $200$ms.

The depth, occlusion, and lighting estimation take two hours to train for a 30-minute long video, while the shadow network takes two days to converge on four Nvidia RTX 2080 Ti GPUs. As shown in the \href{https://youtu.be/bYJ_WdnsEbI}{supplementary video}, applying the depth, occlusion, and lighting estimates to newly inserted objects happens in real-time, while the shadow synthesis takes about 150ms on a single Nvidia RTX 2080 Ti GPU.

\section*{B. Scenes}
\begin{figure}[t]
    \centering
    \includegraphics[width=\linewidth]{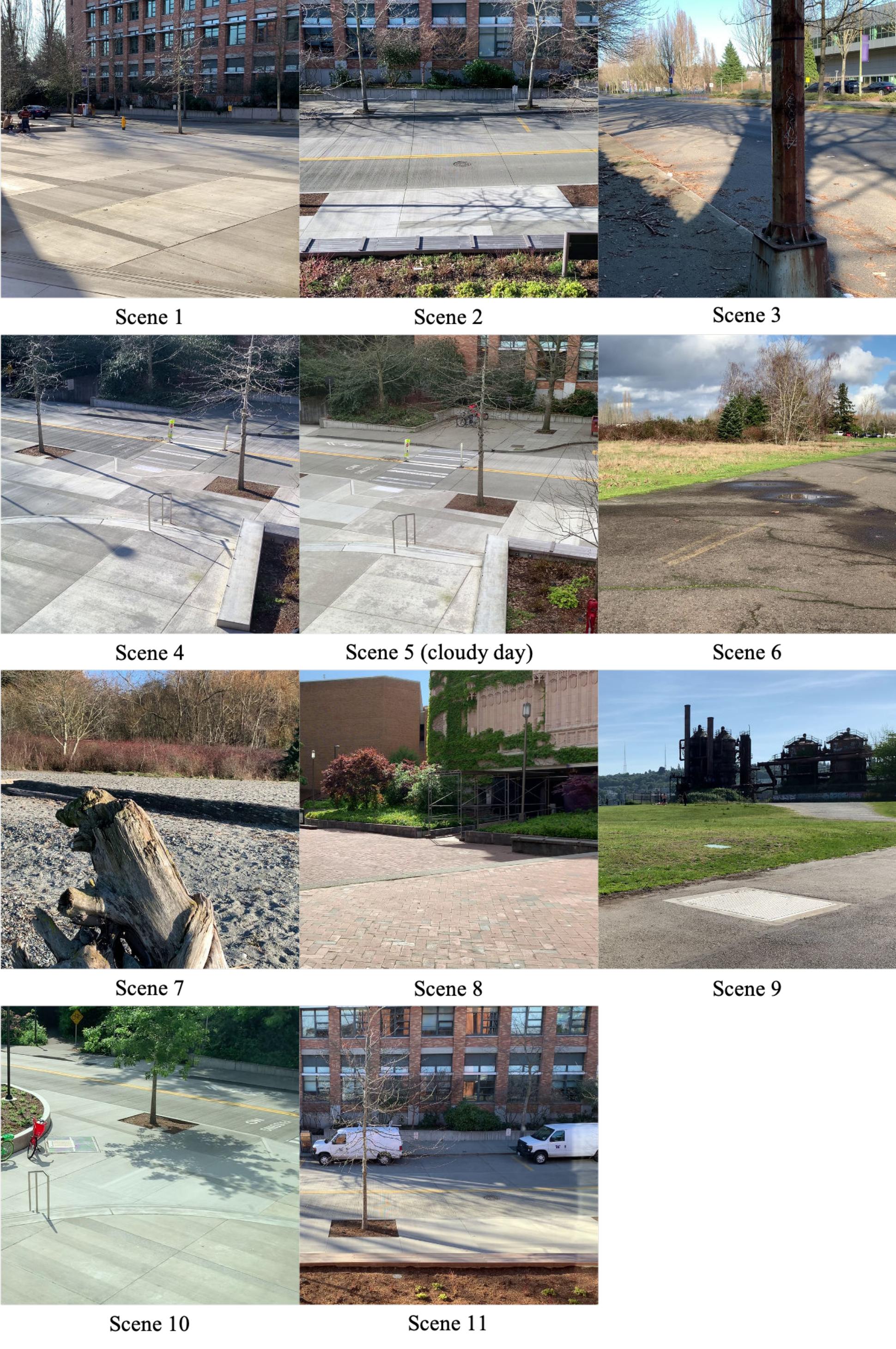}
    \caption{Background images for all the input scenes.}
    \label{fig:scenes}
\end{figure}

We captured 11 video sequences of scenes in a variety of locations including urban settings, parks, and a beach.  Figure~\ref{fig:scenes} shows the background images for those scenes.  Each background image is either an original frame that had no people in it, or, for scenes that were more crowded, the first 1-second median image.

\section*{C. Additional Experiments}

\subsection*{C.1. Importance of $x$-$y$ Channels for Shadow Probing}
\begin{figure}[t]
    \centering
    \includegraphics[width=\linewidth]{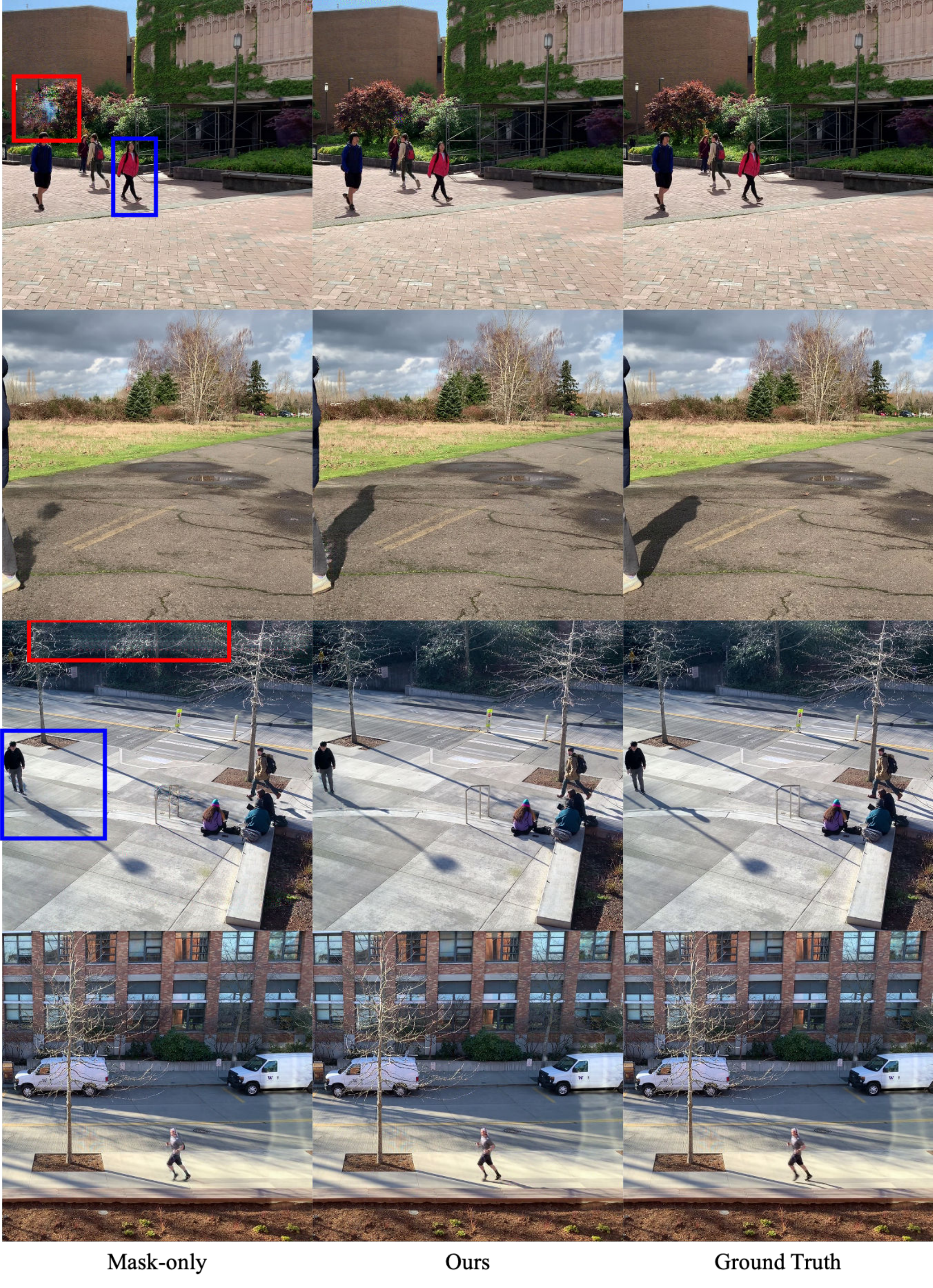}
    \caption{Shadow synthesis on the test set for the mask-only network~(left), our network with mask-$x$-$y$~(middle), and ground truth~(right).  The mask-only network does not learn from explicit $x$-$y$ location information, and shadows do not vary correctly with position, e.g., warping over or being occluded by scene geometry.  Shadow artifacts for mask-only are visible in all scenes; we highlight some with blue boxes to focus on particular inserted people and their shadows.  In addition, the mask-only network generates color patterns, highlighted by red boxes.}
    \label{fig:mask_results}
\end{figure}
To study the importance of the $x$-$y$ channel in our shadow synthesis network (Figure~\ref{fig:network} in the paper), we created a new baseline without it.  I.e., we trained another pix2pix-style network~\cite{wang2018high} that takes only the one-channel mask image and produces bias and gain maps which are applied to the shadow-free image. This mask-only network has the same capacity (5 residual blocks) as our shadow network. Figure~\ref{fig:mask_results} shows example results for shadow estimation on the test set. Without the $x$-$y$ channels, the network fails to learn the geometric impact of the scene, leading to incorrect shadow warping and occlusion.  In addition, the mask-only network tends to produce more repetitive patterns and color artifacts. Our additional $x$-$y$ input helps stabilize the training. 

\subsection*{C.2. Human Study}

To quantify the realism of the synthesized shadows and our shadow synthesis network to baseline pix2pix, we conducted two human studies on (1) the test set (with ground truth reference) and (2) composites with new inserted objects.

For the first study, we collected $75$ test images per scene and generated results with our network and the pix2pix baseline in the paper (shadow-free image $I_{\rm comp}$ as input, direct synthesis of output image with shadow).  Since these were test images, we also had ground truth.  We used Amazon Mechanical Turk (AMT) for the study. We set the requirement for workers to: 1) masters, 2) greater than $97\%$ approval rate.  For each comparison, we showed two images to 3 human subjects, and asked them to choose the one that looked more realistic.  We explicitly told them to focus on the shadow areas, because the inserted object had almost the same appearance.  We set the title of the task to be ``Which image looks more realistic? Focus on the shadows.''.  The order of the image pairs were randomized and each pair was assigned to different workers by AMT.  We took the majority vote over three votes on each image pair.  By taking the majority vote, we reduce the effect of noisy labeling due to ``lazy'' workers who just click randomly.  Overall, our synthesized images were preferred $58.4\%$ of the time to the baseline which was preferred $41.6\%$ of the time (a difference of $16.8\%$). In addition, our results were on average preferred $41.9\%$ of the time to ground truth, whereas the baseline was preferred $34.8\%$ to ground truth. A detailed score breakdown for each scene is shown in Table~\ref{tbl:human}. 

For the second study, we collected 60 image composites: for each scene, 4-7 image pairs -- our shadow composite vs baseline pix2pix (no ground truth available) -- each image with 1-6 inserted objects.  We inserted each random object manually onto the background in locations that made sense (no cars on the sidewalk or people on the street except for crosswalk).  As we had fewer images in the second study, we asked for more workers per comparison in an effort to increase accuracy. In particular, each image pair was shown to 5 human subjects (different workers from the first study), and we again took the majority vote for robustness. Our synthesized images were preferred on average $70.0\%$ of the time to the baseline, demonstrating a clear advantage for novel composites.  Pix2pix struggles significantly more to generalize when inserting new objects that are less similar in appearance to the ones seen during training. A detailed score breakdown for each scene is shown in Table~\ref{tbl:human_insert}. 

\begin{table}[t!]
\centering
\begin{tabular}{l|cccccc}
                     & scene 1 & scene 2 & scene 3 & scene 4  & scene 5  & scene 6 \\
\hline
Ours to baseline     & 61.3   & 50.0   & 65.3   & 49.3    & 62.7    & 58.0   \\
GT to ours           & 86.7   & 54.7   & 64.7   & 54.0    & 51.3    & 54.0   \\
GT to baseline       & 95.3   & 58.7   & 81.3   & 58.0    & 57.3    & 58.7   \\
\hline
                     & scene 7 & scene 8 & scene 9 & scene 10 & scene 11 &         \\
\hline
Ours to baseline     & 58.7   & 58.7   & 51.3   & 68.0    & 58.7    &         \\
GT to ours           & 50.7   & 63.3   & 52.7   & 62.7    & 44.0    &         \\
GT to baseline       & 53.3   & 74.0   & 64.7   & 64.7    & 51.3    &        
\end{tabular}
\caption{Human study results on each scene's test set.}
\label{tbl:human}
\end{table}

\begin{table}[t!]
\centering
\begin{tabular}{l|cccccc}
                 & scene 1 & scene 2 & scene 3 & scene 4  & scene 5  & scene 6 \\
\hline
Ours to baseline & 50.0    & 75.0    & 100.0   & 42.9     & 70.0     & 80.0    \\
\hline
                 & scene 7 & scene 8 & scene 9 & scene 10 & scene 11 &         \\
\hline
Ours to baseline & 70.0    & 75.0    & 75.0    & 70.0     & 78.6     &        
\end{tabular}
\caption{Human study results on inserting new objects for each scene.}
\label{tbl:human_insert}
\end{table}

\subsection*{C.3. Additional Results on Inserting New Objects}
\begin{figure}[t]
    \centering
    \includegraphics[width=\textwidth]{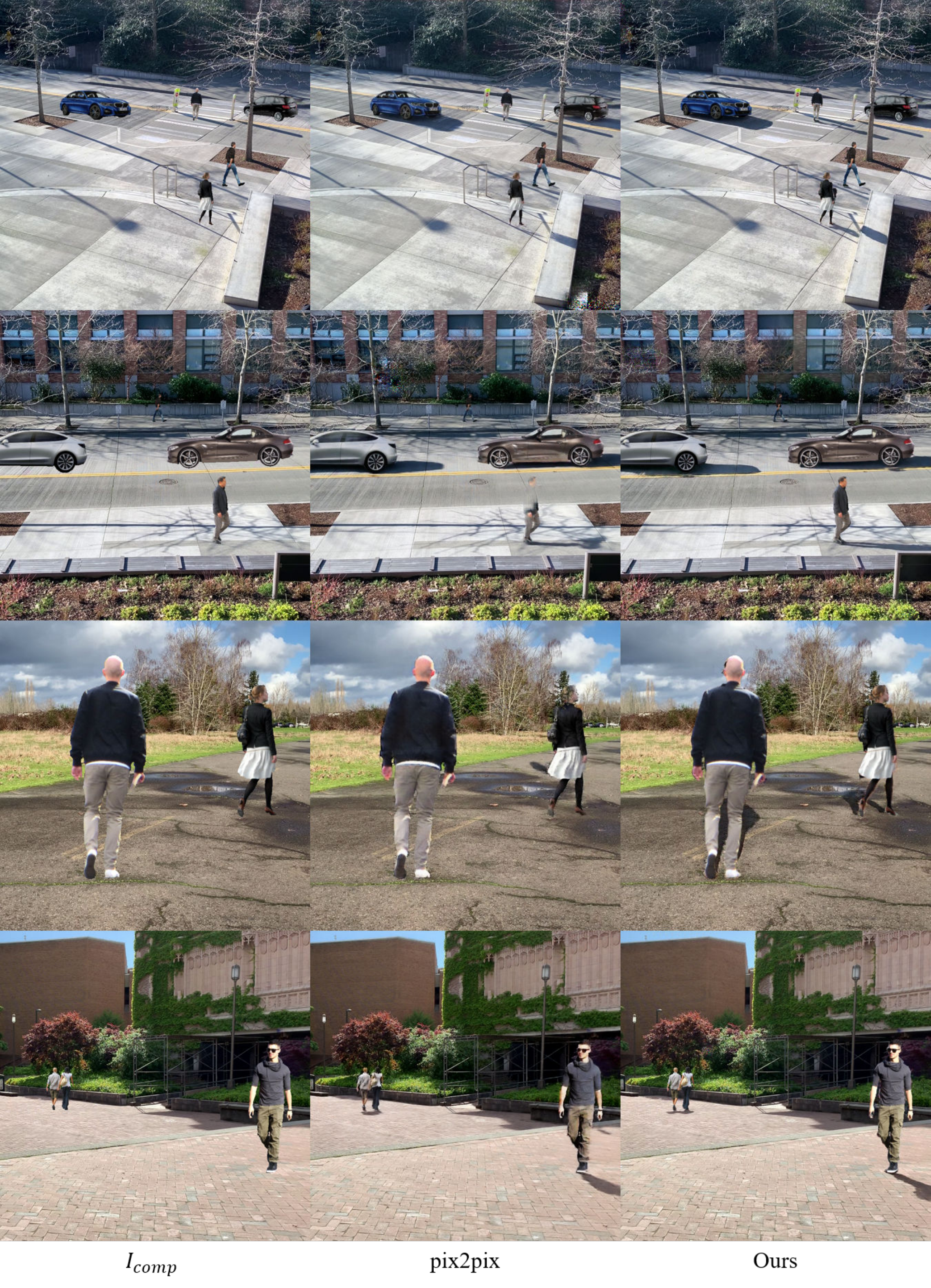}
    \caption{More qualitative results on inserting new objects, with comparison of our shadow network to the pix2pix baseline described in the paper.}
    \label{fig:insert_results_1}
\end{figure}

\begin{figure}[t]
    \centering
    \includegraphics[width=\textwidth]{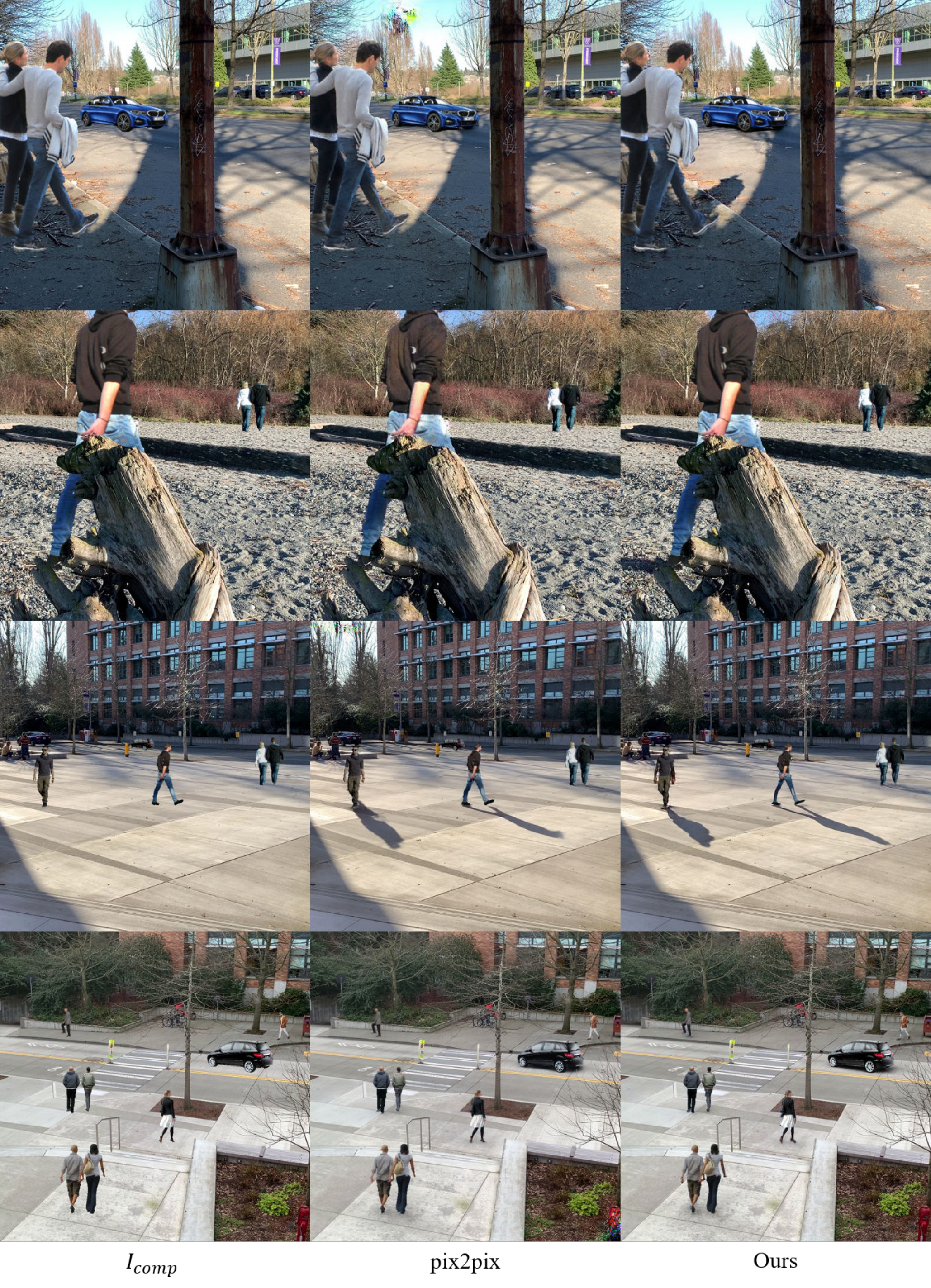}
    \caption{More qualitative results on inserting new objects, with comparison of our shadow network to the pix2pix baseline described in the paper.}
    \label{fig:insert_results_2}
\end{figure}
In Figure~\ref{fig:insert_results_1} and Figure~\ref{fig:insert_results_2}, we show more qualitative results on inserting new objects, with comparison to the pix2pix baseline for the shadow compositing. Our method produces crisper shadows whereas the pix2pix method sometimes produces no shadow at all, generalizing poorly on some of the newly inserted objects and sometimes injecting undesirable color patterns.  For the simpler case of inserting shadows on a cloudy day -- mostly a darkening under/near a person's feet or under a car -- pix2pix performs about as well as our method.  Note that the comparison to pix2pix is just to understand the benefit of our shadow network; both our method and pix2pix benefit from all the other components of our approach (occlusion, lighting, and depth probing).

\section*{D. Supplementary video}
The \href{https://youtu.be/bYJ_WdnsEbI}{supplementary video} demonstrates our compositing tool and shows the importance of each of depth, occlusion, lighting, and shadow probing.

The video also shows that occlusion probing currently does not work well with very thin structures like skinny (leafless) tree branches. This limitation is because all the operations in our method is at pixel level, while these thin structures are in some cases barely a pixel wide.  Furthermore, these thin structures like branches often wiggle in the wind, which can lead to errors in occlusion labeling with our method.

\end{document}